\documentclass{bmvc2k}
\usepackage{amsfonts}
\usepackage{graphicx}
\usepackage{caption}
\usepackage{booktabs}
\usepackage{wrapfig}
\usepackage{soul}
\title{Spherical Vision Transformer for 360$^{\circ}$ Video Saliency Prediction}

\addauthor{Mert Cokelek}{mcokelek21@ku.edu.tr}{1}
\addauthor{Nevrez Imamoglu}{https://nevrez.github.io}{2}
\addauthor{Cagri Ozcinar}{https://cagriozcinar.netlify.app}{3}
\addauthor{Erkut Erdem}{https://web.cs.hacettepe.edu.tr/~erkut}{4}
\addauthor{Aykut Erdem}{https://aykuterdem.github.io}{1}

\addinstitution{KUIS AI Center, Koç University\\ Istanbul, TR}
\addinstitution{AIST\\Tokyo, JP}
\addinstitution{MSK.AI\\ London, UK}
\addinstitution{Hacettepe University\\Istanbul, TR}

\runninghead{Cokelek, Imamoglu, Ozcinar, Erdem, Erdem}{SalViT360}

\definecolor{bostonuniversityred}{rgb}{0.8, 0.0, 0.0}
\definecolor{darkspringgreen}{rgb}{0.01, 0.75, 0.24}

\newcommand{\name}{SalViT360}

\begin{document}
\maketitle
\begin{abstract}
The growing interest in omnidirectional videos (ODVs) that capture the full field-of-view (FOV) has gained 360$^\circ$ saliency prediction importance in computer vision. However, predicting where humans look in 360$^\circ$ scenes presents unique challenges, including spherical distortion, high resolution, and limited labelled data. We propose a novel vision-transformer-based model for omnidirectional videos named \name~that leverages tangent image representations. We introduce a spherical geometry-aware spatio-temporal self-attention mechanism that is capable of effective omnidirectional video understanding. Furthermore, we present a consistency-based unsupervised regularization term for projection-based 360$^{\circ}$ dense-prediction models to reduce artefacts in the predictions that occur after inverse projection. Our approach is the first to employ tangent images for omnidirectional saliency prediction, and our experimental results on three ODV saliency datasets demonstrate its effectiveness compared to the state-of-the-art.

\end{abstract}

\section{Introduction}
\label{sec:intro}
As an important computer vision task, visual saliency prediction aims at predicting where people look in a scene. It is widely used in various areas, such as \textit{saliency-guided} image and video compression~\cite{7962230, zhu2018spatiotemporal, mishra2021multi, 8692638, patney2016towards}, super-resolution~\cite{wang2020visual, hou2022multi}, and quality assessment~\cite{yang2019sgdnet, zhu2021saliency, guan2017visual, 8572733, qiu2021blind} to exploit human perceptual features for enhancement. With the growing popularity of virtual reality (VR) applications and multimedia streaming, predicting saliency in $360^\circ$ videos has received more attention recently.
One primary challenge in processing $360^\circ$ scenes is effectively representing omnidirectional data. Equirectangular Projection (ERP), where the full-FOV scene is projected on a 2D plane, is a common representation due to its computational simplicity. However, ERP suffers from spherical distortion, particularly towards the poles, which can significantly affect the geometric structure of the scene and degrade model performance. While previous works have proposed kernel transformations~\cite{su2017learning} and spherical convolutions~\cite{coors2018spherenet, esteves2018learning} to minimize this distortion on ERP, these methods come at the cost of computational complexity and the loss of global context in $360^\circ$ scenes. Cubemap projection~\cite{4056759} is another common approach that addresses the distortion problem by approximately expressing the spherical scene with six faces of a cube. Although this approach eliminates the distortion to an extent, it breaks the continuity of neighbouring faces and introduces discrepancies in the predictions around the edges.

Previous works in $360^\circ$ saliency prediction primarily focused on addressing this representation problem, with each method trying to balance representative power and computational complexity. Chao~et~al.~\cite{chao2018salgan360} employed cubemap projection and fine-tuned the 2D image saliency model SalGAN~\cite{Pan_2017_SalGAN} on each cube face. Cheng~et~al.~\cite{cheng2018cube} extended cubemap projection with cube-padding to address the discontinuities on the face boundaries. Chao~et~al.~\cite{9122430} extended SalGAN360 with multi-view fusion. Dahou~et~al.~\cite{dahou2021atsal} proposed a two-stream architecture to compute global and local saliency in omnidirectional videos. Their approach uses global prediction as a rough attention estimate, and the local stream on cube faces predicts local saliency. Zhang~et~al.~\cite{Zhang_2018_ECCV} proposed spherical convolutions for saliency prediction. Qiao~et~al.~\cite{qiao2020viewport} showed that the eye fixation distribution bias depends on the viewport locations, which motivated us to introduce {\it spherical position information} into our model. Yun~et~al~\cite{yun2022panoramic} use local undistorted patches with deformable CNNs and use a ViT variant for self-attention across space and time. 
Djilali~et~al.~\cite{Djilali_2021_ICCV} used a self-supervised pre-training based on learning the association between several different views of the same scene and trained a supervised decoder for $360^\circ$ saliency prediction as a downstream task. Although their approach considers the global relationship between viewports, it ignores the temporal dimension that is crucial for video understanding.

The methods mentioned above share a common limitation in processing $360^\circ$ data through projections or modified kernels and ignore the full field-of-view, which is critical for global scene understanding and saliency prediction. Thus, there is a need for an effective omnidirectional data processing method that minimizes spherical distortion while preserving the global context and avoiding computational overhead and artefacts introduced by the previous methods. Recently, Eder~et~al.~\cite{eder2020tangent} proposed tangent image representations, which use gnomonic projection to map a spherical image into multiple overlapping patches, where each patch is tangent to the faces of an icosahedron. This method tackles the problem of spherical distortion on the scene. However, to our interest, the dense-prediction models particularly suffer from discrepancies and artefacts on overlapping regions of tangent image patches after inverse projection to ERP. In this work, we propose using tangent images to process undistorted local viewports and develop a transformer-based model to learn their global association for saliency prediction in 360$^\circ$ videos. This is the first work that employs tangent images for omnidirectional saliency prediction, which is also capable of modelling the temporal dimension,  motivated by the recent video transformer architectures, e.g., ~\cite{gberta_2021_ICML}.

\textbf{Proposed method and contributions:} We use gnomonic projection to obtain multiple undistorted tangent images, which enables us to extract rich local spatial features using any pre-trained and fixed 2D backbone. Our extensive experiments
demonstrate the effectiveness of our proposed \name~model against the state-of-the-art on VR-EyeTracking~\cite{xu2018gaze}, PVS-HMEM~\cite{xu2018predicting}, and 360-AV-HM~\cite{9301766} datasets. Our contributions are three-fold:
\begin{itemize}
    \item {\it Spherical geometry-aware spatial attention.} We aggregate global information on the sphere by computing self-attention among tangent viewports. We use tangent features as viewport tokens to address the quadratic complexity associated with spatial self-attention. We introduce $360^\circ$ geometry awareness to the transformer by using learnable spherical position embeddings guided by per-pixel angular coordinates $(\phi, \theta)$, denoting \textit{latitude} and \textit{longitude} angles, respectively. The proposed position embedding method outperforms standard 1D embeddings, and it can easily be integrated into any transformer architecture designed for $360^\circ$ images.  

    \item {\it Viewport Spatio-Temporal Attention (VSTA).} The complexity of joint spatio-temporal self-attention increases quadratically with respect to the number of frames. With VSTA, we optimize this joint computation on tangent viewports from consecutive frames, where the spatio-temporal self-attention is performed in two stages: (1) Viewport Spatial Attention (VSA) and (2) Viewport Temporal Attention (VTA). In VSA, spherical geometry-aware self-attention is computed \textit{intra-frame} level. The temporal information in the videos is encoded by a VTA, among tangent planes that point to the same direction in the \textit{inter-frame} level.
    
    \item {\it Viewport Augmentation Consistency (VAC).} We propose an unsupervised, consistency-based loss for omnidirectional images, minimizing the discrepancies in overlapping regions of tangent predictions. The loss is computed between the weight-sharing saliency predictions of two tangent image sets which are generated with different configurations. This regularization method is suitable for any projection-based dense-prediction model. Importantly, VAC does not introduce any memory or time overhead during test time, as only one set of predictions is sufficient for inference.
\end{itemize}

\begin{figure*}[!t]
    \centering
    \includegraphics[width=0.8\textwidth]{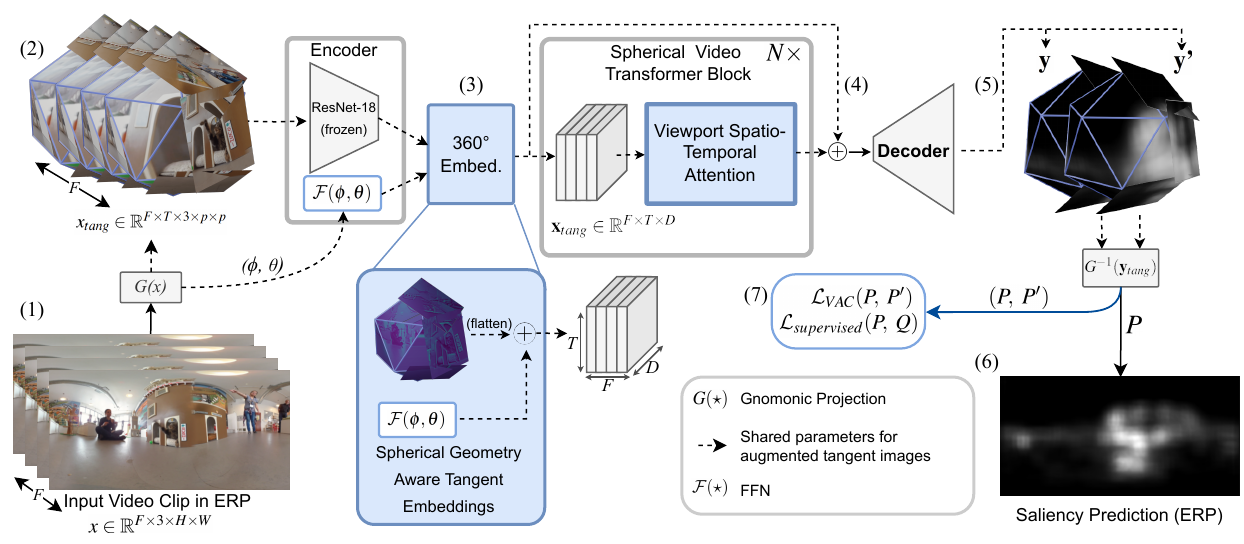}
    \vspace{1mm}
    \caption{\textbf{Overview of the proposed \name~model.} The ERP video clip of $F$ frames (1) is projected to $F\times T$ tangent images per set (2). Each tangent image is encoded and fused with spherical-geometry-aware position embeddings (3) for the 360$^\circ$ video transformer to aggregate global information (4). The outputs are decoded into saliency predictions in tangent space (5), which are projected back to ERP, giving the final saliency map (6). In addition to the supervised loss, the model is trainable with $\mathcal{L}_{VAC}(P,\ P')$ to minimize the tangent artefacts (7). During test time, the model works with a single tangent set. For simplicity, only one set of tangent images is shown.}
    \label{fig:overview}
\end{figure*}

\section{Method}\label{method}
In Fig.~\ref{fig:overview}, we present an overview of our proposed model \name. We start with gnomonic projection~\cite{eder2020tangent} to obtain tangent images for each frame in the input video clip. These are passed to an encoder-transformer-decoder architecture. 
The image encoder extracts local features for each tangent viewport and reduces the input dimension for the subsequent self-attention stage. 
We map the pixel-wise angular coordinates to produce the proposed spherical geometry-aware position embeddings $\mathcal{F}(\phi,\theta)$ for the 360$^\circ$ transformer, enabling better learning of spatial representations. 
The transformer utilizes Viewport Spatio-Temporal Attention (VSTA) to capture inter and intra-frame global information across tangent viewports in a temporal window. 
The transformed embeddings are then fed into a 2D CNN-based decoder, which predicts saliency on the tangent images. We then apply inverse gnomonic projection on the tangent predictions to obtain the final saliency maps in ERP. We propose an unsupervised consistency-based Viewport Augmentation Consistency Loss to mitigate the discrepancies after inverse gnomonic projection. The learnable parameters of the network are in tangent space, allowing us to use large-scale pre-trained 2D models for feature extraction, while the rest of the network is trained from scratch.

\vspace{3mm}\noindent\textbf{Gnomonic Projection and Encoder.} We first project the input ERP clip $x\in\mathbb{R}^{F\times3\times{H}\times{W}}$ to a set of tangent clips $x_{tang}\in\mathbb{R}^{F\times{T}\times3\times{p}\times{p}}$, where $F$, $3$, $H$, and $W$ are the number of frames, channel dimension (RGB), height, and width of a given video, respectively. The resulting tangent images have a patch size of $p\times p=224\times224$ pixels. 
Number of tangent images per frame $T$, and {\it FOV} are the projection hyperparameters which vary between  $10/18$ and $120^\circ/80^\circ$.
We downsample and flatten the encoder features to obtain tangent feature vectors with dimension $D=512$. We map the angular coordinates $(\phi, \theta)$ for each pixel of the tangent viewports to the same feature dimension using an FC layer and sum these embeddings with encoder features to obtain the proposed \textit{spherical geometry-aware embeddings}~$\mathbf{x}_{tang}\in\mathbb{R}^{{F}\times{T}\times{D}}$ that are used in the transformer. 

\begin{wrapfigure}{r}
{0.5\textwidth}
\vspace{-4mm}%
\centering
\includegraphics[width=0.87\linewidth]{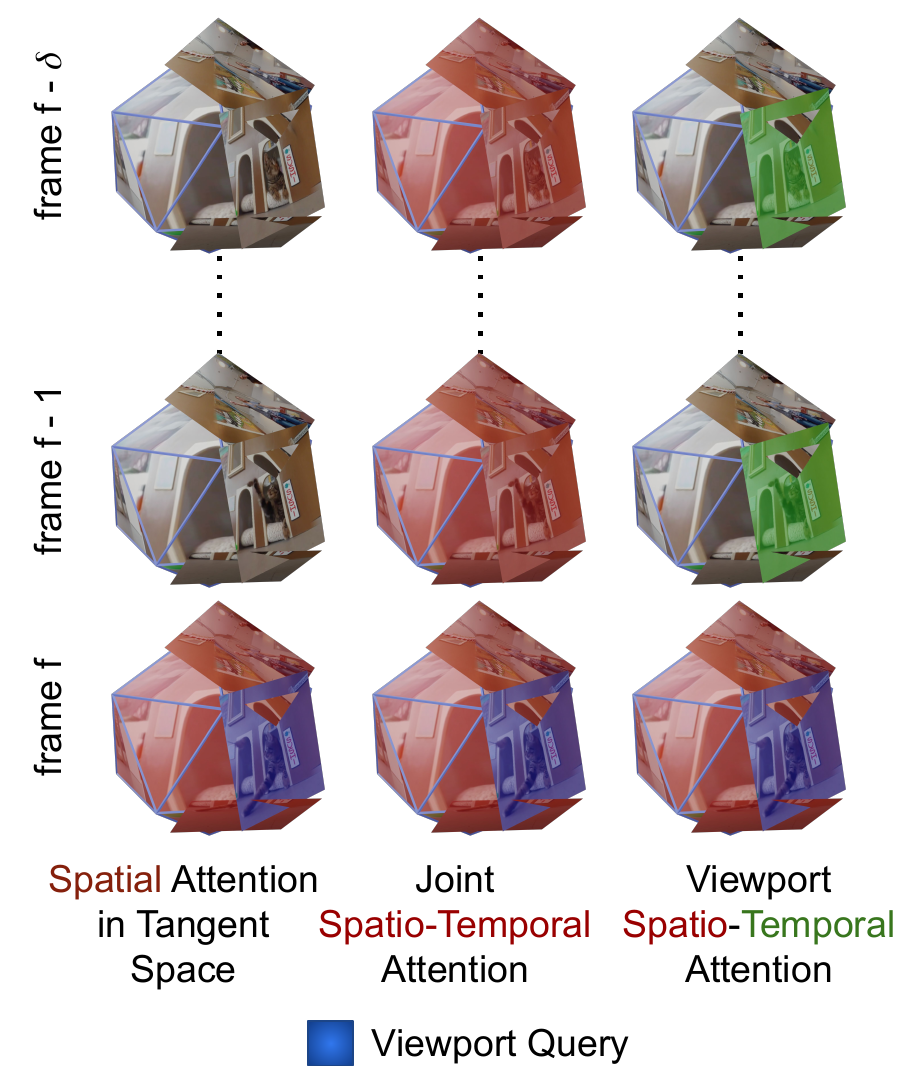}
\caption{\textbf{The proposed Viewport Spatio-Temporal Attention (VSTA)} (\textit{right}), as compared to Viewport Spatial Attention (VSA) (\textit{left}) and Joint Spatio-Temporal Attention (\textit{middle}). {\color{bostonuniversityred}Red} and {\color{darkspringgreen}Green} viewports denote the self-attention neighborhood for each scheme.}
\vspace{-2.5mm}
\label{fig:attention}
\end{wrapfigure}

\vspace{3mm}\noindent\textbf{Viewport Spatio-Temporal Attention for 360$^\circ$ videos.} While the pre-trained encoder extracts rich spatial features for each tangent image locally, aggregating the global context in the full-FOV is essential for 360$^\circ$ scene understanding. We propose a self-attention mechanism on tangent viewport features to achieve this. However, since incorporating the temporal dimension of the videos increases the number of tokens and thus the computational complexity, we approximate spatio-temporal attention with two stages: we apply temporal attention (1) among the same tangent viewports from consecutive $F$ frames, then, spatial attention (2) among $T$ tangent viewports in the same frame. This reduces the overall self-attention complexity from $F^2\times{T^2}$ to $F^2 + T^2$. In this way, we effectively model the global context required for 360$^\circ$ video understanding. Fig.~\ref{fig:attention} illustrates our proposed \textit{Viewport Spatio-Temporal Attention} for 360$^\circ$ videos.

\vspace{3mm}\noindent\textbf{Decoder.} The decoder comprises four upsample layers followed by $3\times3$ convolutions and normalization layers. For a set of tangent clips, it takes the skip connection of encoder and transformer features $\mathbf{x}_{tang}^f\in\mathbb{R}^{T\times D\times7\times7}$ of the last frame as input and outputs saliency prediction $\mathbf{y}\in\mathbb{R}^{T\times56\times56}$ on tangent planes. The final ERP saliency maps are obtained by passing the tangent predictions to inverse gnomonic projection. We aggregate global information among tangent images through the transformer, however, each tangent plane is predicted separately in the decoder. %

\vspace{3mm}\noindent\textbf{Viewport Augmentation Consistency (VAC).} Our model effectively learns the overall saliency distribution with the supervised loss. However, since each tangent plane is predicted separately, the final ERP saliency map contains discrepancies in overlapping regions of tangent patches. To tackle this issue, we propose an unsupervised loss strategy, called \textit{Viewport Augmentation Consistency} (VAC), for improving the consistency between the saliency predictions $P$ and $P'$ from two tangent projection sets. Specifically, we generate the second tangent set by applying different configurations, such as horizontally shifting the tangent planes on the sphere, using a larger FOV for the same viewports, and varying the number of tangent images at different viewports. We provide a detailed comparison of these approaches in the supplementary. VAC does not require any additional memory or time overhead since it uses the shared parameters of the whole model, and the forward pass is done in parallel. Furthermore, since the ERP predictions from the original $P$ and augmented $P'$ tangent sets are expected to be consistent, only one tangent set is sufficient for testing. The VAC loss is defined as:
\begin{equation}\label{eqn:VACLoss}
\begin{aligned}
    \mathcal{L}_{VAC}(P,\ P') &= \mathcal L^{weighted}_{KLD}(P,\ P')\ +\ \mathcal L^{weighted}_{CC}(P,\ P'),\\
     \mathcal{L}^{weighted}_{KLD}(P,\ P') &= \sum_{\substack{i, j}} {P_{i, j}\log{(\epsilon+\frac{P_{i, j}}{P'_{i, j}+\epsilon}})}\cdot {w}_{i, j}, \\
     \mathcal{L}^{weighted}_{CC}(P,\ P') &= 1 -\frac{\sum(P \cdot P')\cdot w_{i,j}} {\sum(P \cdot P) \cdot\ \sum(P' \cdot P')}
\end{aligned}
\end{equation}
where $P,\ P'$ are the saliency predictions from original and augmented viewports, and ${w}$ is an optional weight matrix obtained from gnomonic projection to weigh the overlapping pixels of gnomonic projection on ERP predictions. Details for viewport augmentation approaches and the weighting operation are provided in the supplementary.

\section{Experiments}
\subsection{Setup}
\vspace{3mm}\noindent\textbf{Datasets and pre-processing.} We use the publicly available \textit{VR-EyeTracking}~\cite{xu2018gaze} dataset for training, which consists of $134$ train and $74$ test videos viewed by at least $31$ subjects, lasting between $20-60$ seconds. We sampled the videos at 16 fps with a resolution of $960\times1920$. For cross-dataset evaluation, we use the \textit{PVS-HMEM}~\cite{xu2018predicting} and \textit{360AV-HM}~\cite{9301766} datasets, which respectively contain $76$ and $21$ videos viewed by $58$ and $15$ subjects. The videos in the \textit{PVS-HMEM} dataset have varying durations between $10-80$ secs, while those in the \textit{360AV-HM} dataset have a duration of 25 secs. The videos in both datasets have a frame rate between 24-60 fps.

\vspace{3mm}\noindent\textbf{Evaluation metrics and loss functions.} We evaluate the performance of the models using the four most commonly used saliency evaluation metrics~\cite{bylinskii2018different}: Normalized Scanpath Saliency (NSS), KL-Divergence (KLD), Correlation Coefficient (CC), and Similarity Metric (SIM). We use a weighted differentiable combination of KLD, CC, and Selective-MSE (MSE on normalized saliency maps at only eye-fixation points~\cite{salfbnet}) for the supervised loss, as given below:
\begin{equation}\label{loss}
{\mathcal{L}_{supervised}(P,\ {Q}_{s},\ {Q}_{f}) = \mathcal L_{KLD}(P,\ {Q}_s)\ +\ \mathcal L_{CC}(P,\ Q_s)\ +\ \alpha\mathcal L_{SMSE}(P,\ {Q}_s,\ {Q}_f)}
\end{equation}
where $P,\ {Q}_s,\ {Q}_f$ are the predicted saliency, ground truth density and fixation maps, respectively, and $\alpha = 0.005$.

\vspace{3mm}\noindent\textbf{Architecture and optimization details.} We use a ResNet-18 encoder pre-trained on ImageNet and keep it frozen while extracting features. Our baseline 360$^\circ$ video transformer consists of 6 blocks with an embedding dimension of 512 and 8 attention heads. For the spatial position embeddings, we use linear projections of flattened pixel-wise angular coordinates, while we use rotary embeddings are used for the time dimension with a window size of $F=8$. As in~\cite{gberta_2021_ICML}, we apply temporal self-attention followed by spatial self-attention in alternation. We train our \name~model using the AdamW optimizer~\cite{loshchilov2017decoupled} with an initial learning rate of $1e-5$, default weight decay, and momentum parameters of $1e-2$ and $(\beta_1,\ \beta_2)=(0.9,\ 0.999)$, with a batch size of 16 for five epochs with early stopping.%
\begin{figure*}[!t]
    \centering
    \includegraphics[trim= 5 0 0 0,clip,width=0.99\textwidth]{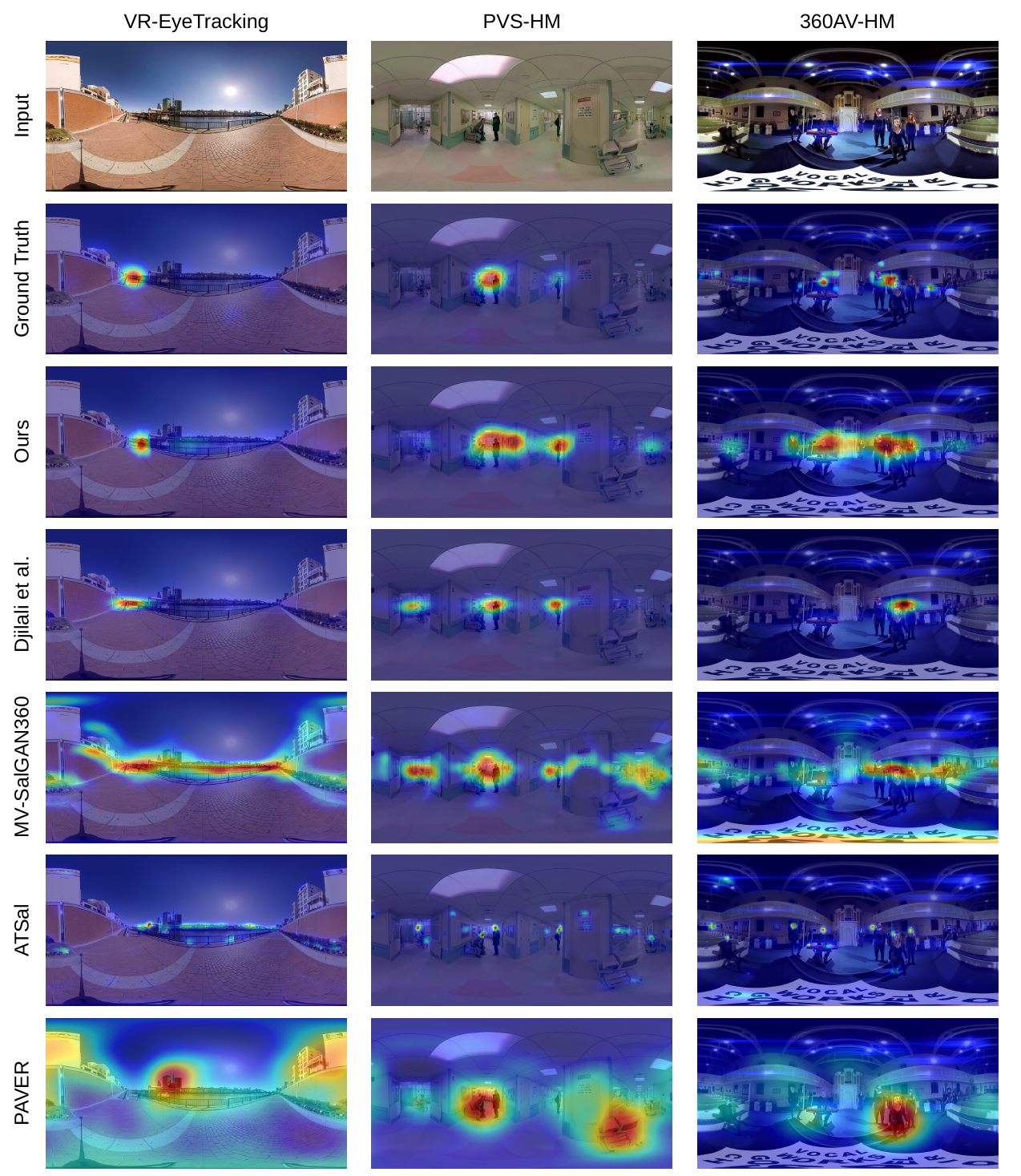}
    \vspace{3mm}
    \caption{\textbf{Qualitative comparison on VR-EyeTracking~\cite{xu2018gaze}, PVS-HM~\cite{xu2018predicting}, and 360AV-HM~\cite{9301766} datasets.} Our proposed approach gives better results compared to existing models. The saliency predictions of our model better resemble the ground truth fixation density maps, producing sparse estimates while covering the most dominant modes better.
    }
    \label{fig:qualitative}
\end{figure*}

\begin{table*}[!t]
    \centering
    \caption{\textbf{Performance analysis of \textit{\name}~against the state-of-the-art 360$^{\circ}$ saliency models} on VR-EyeTracking, PVS-HMEM, 360AV-HM datasets. While the scores in \textbf{bold} highlight the best performance, the \underline{underlined} ones are the second best.}
    \vspace{0.025\linewidth}
    \resizebox{\linewidth}{!}{
    \begin{tabular}{lcccccccccccc}
        \toprule
        & \multicolumn{4}{c}{{\bf VR-EyeTracking}} & \multicolumn{4}{c}{{\bf PVS-HMEM}} & \multicolumn{4}{c}{{\bf 360AV-HM}} \\
        \cmidrule(lr){2-5} \cmidrule(lr){6-9} \cmidrule(lr){10-13}
        {\bf Method} & {\bf NSS$\uparrow$} & {\bf KLD$\downarrow$} & {\bf CC$\uparrow$} & {\bf SIM$\uparrow$} & {\bf NSS$\uparrow$} & {\bf KLD$\downarrow$} & {\bf CC$\uparrow$} & {\bf SIM$\uparrow$} & {\bf NSS$\uparrow$} & {\bf KLD$\downarrow$} & {\bf CC$\uparrow$} & {\bf SIM$\uparrow$} \\
        \midrule
        CP-360              & 0.624 & 15.338 & 0.165 & 0.240 & 0.576 & 4.738 & 0.162 & 0.198 & 0.689 & 24.426 & 0.061 & 0.041 \\
        SalGAN360           & 1.753 & 10.845 & 0.370 & 0.355 & 1.513 & 4.394 & 0.314 & 0.291 & 0.719 & 25.301 & 0.065 & 0.036 \\
        MV-SalGAN360        & 1.818 & 8.713  & 0.382 & 0.357 & 1.546 & 4.112 & 0.316 & 0.295 & 0.716 & 25.322 & 0.066 & 0.036 \\
        ATSal               & 1.317 & 12.259 & 0.336 & 0.318 & 0.732 & 4.303 & 0.183 & 0.219 & 0.727 & 24.141 & 0.058 & 0.041 \\
        PAVER               & 1.511 & 13.267 & 0.307 & 0.294 & 0.750 & 3.736 & 0.224 & 0.269 & 0.732 & 23.944 & 0.065 & 0.035 \\
        Djilali~et~al.      & \textbf{3.183} & \underline{6.570} & \underline{0.565} & \underline{0.475}  & \underline{1.688} & \underline{2.430} & \underline{0.447} & \underline{0.404} & \underline{1.727} & \underline{22.889} & \underline{0.148} & \underline{0.085} \\
        \midrule
        \textit{\name~(Ours)}  & \underline{2.630} & \textbf{5.744} & \textbf{0.586} & \textbf{0.492} & \textbf{2.191} & \textbf{1.841} & \textbf{0.626} & \textbf{0.495} & \textbf{1.946} & \textbf{22.711} & \textbf{0.168} & \textbf{0.093} \\
        \bottomrule
    \end{tabular}}
    \label{table:sotacomparison}
\end{table*}
\subsection{Comparison with the State-of-the-art}
In Table~\ref{table:sotacomparison}, we present the results of our \name~model and the existing models. We evaluate the performance of \name~with six state-of-the-art models for 360$^\circ$ image and video saliency prediction, namely CP-360~\cite{cheng2018cube}, SalGAN360~\cite{chao2018salgan360}, MV-SalGAN360~\cite{9122430}, Djilali et al.~\cite{Djilali_2021_ICCV}, ATSal~\cite{dahou2021atsal}, PAVER~\cite{yun2022panoramic}. ATSal, PAVER, and CP-360 are video saliency models; the rest are image-based models developed for the omnidirectional domain. On VR-EyeTracking test set, \name~outperforms the state-of-the-art on three metrics and gives competitive results on NSS. On PVS-HMEM and 360-AVHM datasets, it outperforms the state-of-the-art by a margin on all metrics, demonstrating our proposed model's cross-dataset performance. These results demonstrate that \name~has a better generalization capability than the existing methods. The qualitative comparison in Fig.~\ref{fig:qualitative} also shows the effectiveness of our approach in highlighting the salient regions more accurately.

\subsection{Experimental Analysis and Ablation Studies} To assess the contribution of each component of our approach and to provide an in-depth analysis of spatio-temporal modelling, we perform additional experiments on the test split of VR-EyeTracking dataset and report our findings in Table~\ref{table:ablation} and Table~\ref{table:stmodelling}. In these experiments, we consider a baseline model comprised of a 2D ResNet-18 backbone, a vision transformer with Viewport Spatial Attention, and a CNN-based decoder.

\vspace{3mm}\noindent\textbf{Spherical Position Embeddings.} We compare the performance of our proposed \textit{spherical geometry-aware spatial position embeddings} with regular 1D learnable position embeddings. The results on all four metrics show that our proposed embedding method outperforms it, demonstrating that it is more suitable for processing spherical data with Vision Transformers.

\vspace{2mm}\noindent\textbf{Viewport Augmentation Consistency and Late-Fusion.} We train our VSTA baseline using only the supervised loss. We then compare this single-scale baseline to the one trained with a weighted combination of the supervised loss ($\mathcal{L}_{supervised}$) and the consistency loss ($\mathcal{L}_{VAC}$). Table~\ref{table:ablation} shows that VAC outperforms the VSTA baseline on three distribution-based saliency evaluation metrics with a performance gain of $3.2\%$ on KLD, $6.4\%$ on CC, and $5.8\%$ on SIM, respectively. Lastly, we investigated the effect of using the predictions of two tangent sets in the final prediction. We perform this with an optional late-fusion as element-wise multiplication of two ERP predictions to highlight the consistently predicted salient regions better. This simple optional fusion improves the performance on three metrics significantly, with zero memory- and $0.5\times$ time-overhead. We provide sample results for the qualitative comparison of these components in~Fig.~\ref{fig:ablation}, and detail them in the supplementary.

\begin{table}[t!]
    \centering
    \caption{\textbf{Ablation study} for each component in our approach on the test split of the VR-EyeTracking dataset.} 
    \vspace{0.025\linewidth}
    \begin{tabular}{lcccccc}
        \toprule
        {\bf Method} & {\bf \#\ params} & {\bf NSS$\uparrow$} & {\bf KLD$\downarrow$} & {\bf CC$\uparrow$} & {\bf SIM$\uparrow$}  \\
        \midrule
        VSA $(w/\ {1D\ Pos.\ Emb.})$                    & 30.28M & 2.518 & 6.445 & 0.560 & 0.472 \\
        {\ \ \ \ }$+\ Spherical\ Pos.\ Emb.$            & 30.78M & 2.575 & 6.221 & 0.563 & 0.475 \\
        VSTA $(w/\ {Sph. \ Pos.\ Emb.})$                   & 37.07M & \textbf{2.664} & 6.174 & 0.570 & 0.479 \\
        {\ \ \ \ } $\ +\ VAC\ (w/o\ mask)$              & 37.07M & {2.624} & {6.011} & {0.576} & {0.490} \\
        {\ \ \ \ } $\ +\ VAC\ (w/\ mask)$                & 37.07M & \underline{2.630} & \underline{5.744} & \underline{0.586} & \underline{0.492} \\
        {\ \ \ \ } $\ +\ $ {\it Late-Fusion}             & 37.07M & {2.578} & \textbf{4.654} & \textbf{0.592} & \textbf{0.495} \\
        \bottomrule
    \end{tabular}
    \label{table:ablation}
\end{table}

\begin{table}[!b]
    \centering
    \caption{\textbf{Spatio-temporal modelling performance} of \textit{\name}~compared against two alternative approaches on the test split of the VR-EyeTracking dataset.} 
    \vspace{0.025\linewidth}
    \begin{tabular}{lcccccc}
        \toprule
        {\bf Method} & {\bf \#\ params} & {\bf NSS$\uparrow$} & {\bf KLD$\downarrow$} & {\bf CC$\uparrow$} & {\bf SIM$\uparrow$}  \\
        \midrule
        VSA $\ +\ $ \textit{2+1D-CNN Enc.}                    & 28.82M & 2.568 & 5.915 & 0.568 & 0.477 \\
        VSA $\ +\ $ \textit{Offline EMA}                         & 30.78M & 2.591 & 6.018 & 0.566 & 0.477 \\
        \textit{\name}~(VSTA $\ +\ $ VAC)                                  & 37.07M & \textbf{2.630} & \textbf{5.744} & \textbf{0.586} & \textbf{0.492} \\
        \bottomrule
    \end{tabular}
    \label{table:stmodelling}
\end{table}

\vspace{3mm}\noindent\textbf{Spatio-Temporal modelling.} We conducted several experiments to evaluate the effectiveness of our proposed Viewport Spatio-Temporal Attention (VSTA) mechanism. In Table~\ref{table:ablation}, we compare Viewport Spatial Attention (VSA) and VSTA blocks to assess the contribution of temporal information processing in omnidirectional videos. In Table~\ref{table:stmodelling}, we compare our VSTA with two distinct approaches, namely {\it 2+1D-CNN}~\cite{morgado2020learning} backbone and \textit{Offline EMA}. {\it 2+1D-CNN} backbone is an R2+1D model~\cite{8578773} which performs convolution over consecutive frames, pre-trained on undistorted \textit{normal-FOV} crops in $360^\circ$ videos. We replace ResNet-18 + VSTA with {\it 2+1D-CNN} + VSA to introduce temporal features for spatial self-attention. In the other setting, we keep ResNet-18 and VSA and apply a weighted exponential moving average on $F$ consecutive predictions for temporal aggregation. Additionally, we ablate on transformer depth, which shows that our VSTA blocks gradually learn better spatio-temporal representations in deeper layers. Our experiments demonstrate that the proposed VSTA mechanism outperforms the spatial-only setting and the other two spatio-temporal approaches. We refer the reader to the supplementary for comprehensive experiments on the effect of {\it the transformer depth,} and {\it temporal window size $F$}, along with the performance of {\it joint spatio-temporal attention.}

\begin{figure*}[!t]
    \centering
    \includegraphics[trim= 0 0 0 0,clip,width=0.99\textwidth]{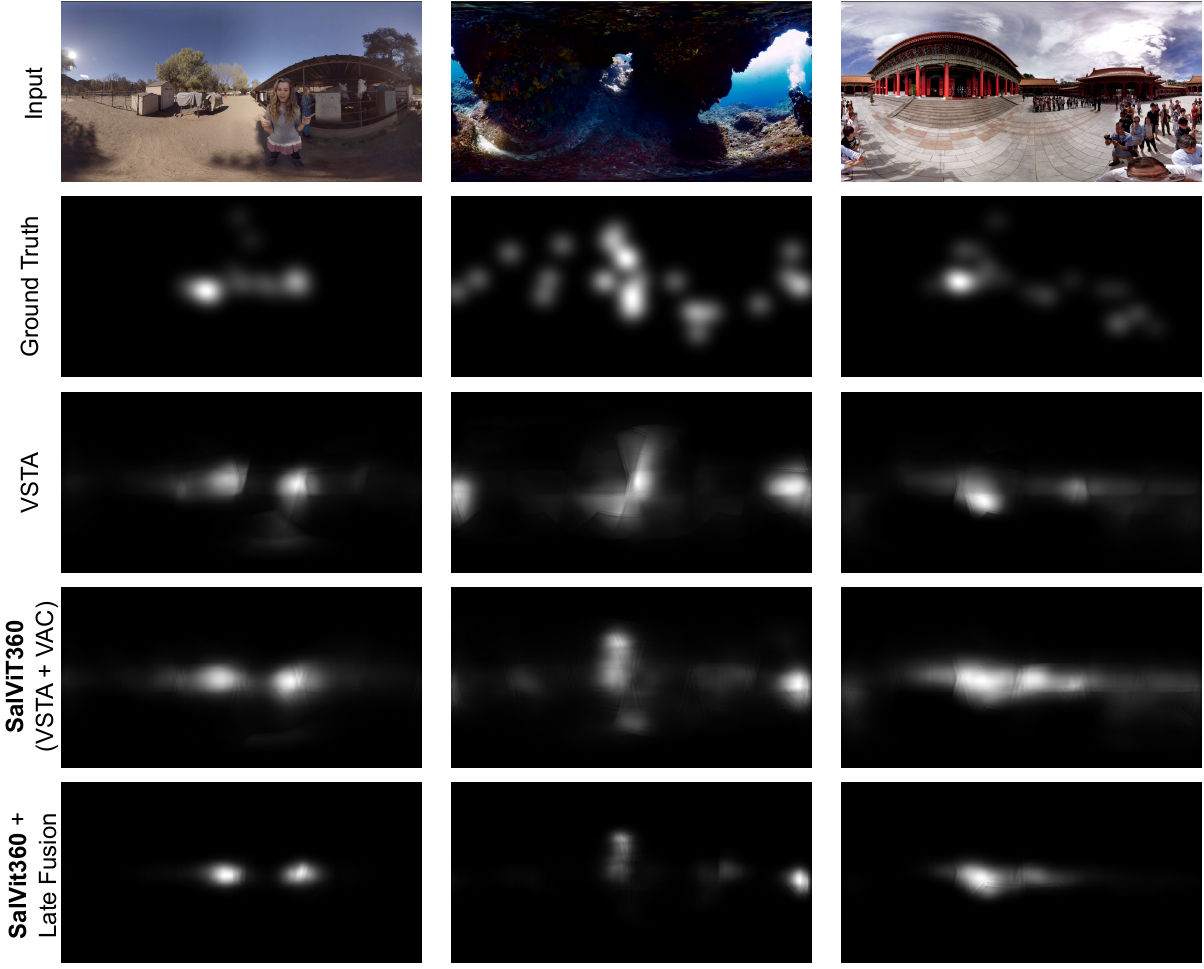}
    \vspace{3mm}
    \caption{\textbf{Qualitative comparison} for our \textit{VSTA baseline} \textit{(third row)}, with the proposed \textit{VAC loss (fourth row)}, and the optional \textit{late-fusion (last row)}, compared to the ground truth.}  
    \label{fig:ablation}
\end{figure*}

\section{Conclusion}
In this study, we proposed \name, a transformer-based framework using tangent image representations for 360$^\circ$ video saliency prediction. We also introduced a spatio-temporal attention mechanism on tangent viewports to effectively capture the global and temporal context in omnidirectional videos. Our framework employs a new (model-agnostic) spherical geometry-aware position embedding structure based on angular coordinates. Lastly, we suggested an unsupervised, consistency-based loss function as a regularizer to the artefacts commonly observed in projection-based dense-prediction models. Our experimental results obtained from three omnidirectional video saliency datasets demonstrate that our proposed \name~model outperforms the state-of-the-art qualitatively and quantitatively. As a future research direction, we plan to investigate the impact of spatial audio in 360$^\circ$ videos and extend our proposal to the audio-visual domain.
\bibliography{egbib}

\begin{thebibliography}{10}

\bibitem{7962230}
Shengxi Li, Mai Xu, Yun Ren, and Zulin Wang.
\newblock Closed-form optimization on saliency-guided image compression for
  hevc-msp.
\newblock {\em IEEE Transactions on Multimedia}, 20(1):155--170, 2018.

\bibitem{zhu2018spatiotemporal}
Shiping Zhu and Ziyao Xu.
\newblock Spatiotemporal visual saliency guided perceptual high efficiency
  video coding with neural network.
\newblock {\em Neurocomputing}, 275:511--522, 2018.

\bibitem{mishra2021multi}
Dipti Mishra, Satish~Kumar Singh, Rajat~Kumar Singh, and Divanshu Kedia.
\newblock Multi-scale network (mssg-cnn) for joint image and saliency map
  learning-based compression.
\newblock {\em Neurocomputing}, 460:95--105, 2021.

\bibitem{8692638}
Shiping Zhu, Chang Liu, and Ziyao Xu.
\newblock High-definition video compression system based on perception guidance
  of salient information of a convolutional neural network and hevc compression
  domain.
\newblock {\em IEEE Transactions on Circuits and Systems for Video Technology},
  30(7):1946--1959, 2020.

\bibitem{patney2016towards}
Anjul Patney, Marco Salvi, Joohwan Kim, Anton Kaplanyan, Chris Wyman, Nir
  Benty, David Luebke, and Aaron Lefohn.
\newblock Towards foveated rendering for gaze-tracked virtual reality.
\newblock {\em ACM Trans. Graph. (TOG)}, 35(6):1--12, 2016.

\bibitem{wang2020visual}
Haoxiang Wang, Zhihui Li, Yang Li, Brij~B Gupta, and Chang Choi.
\newblock Visual saliency guided complex image retrieval.
\newblock {\em Pattern Recognition Letters}, 130:64--72, 2020.

\bibitem{hou2022multi}
Zejiang Hou and Sun-Yuan Kung.
\newblock Multi-dimensional dynamic model compression for efficient image
  super-resolution.
\newblock In {\em Proceedings of the IEEE/CVF Winter Conference on Applications
  of Computer Vision}, pages 633--643, 2022.

\bibitem{yang2019sgdnet}
Sheng Yang, Qiuping Jiang, Weisi Lin, and Yongtao Wang.
\newblock Sgdnet: An end-to-end saliency-guided deep neural network for
  no-reference image quality assessment.
\newblock In {\em Proceedings of the 27th ACM International Conference on
  Multimedia}, pages 1383--1391, 2019.

\bibitem{zhu2021saliency}
Mengmeng Zhu, Guanqun Hou, Xinjia Chen, Jiaxing Xie, Haixian Lu, and Jun Che.
\newblock Saliency-guided transformer network combined with local embedding for
  no-reference image quality assessment.
\newblock In {\em Proceedings of the IEEE/CVF International Conference on
  Computer Vision}, pages 1953--1962, 2021.

\bibitem{guan2017visual}
Jingwei Guan, Shuai Yi, Xingyu Zeng, Wai-Kuen Cham, and Xiaogang Wang.
\newblock Visual importance and distortion guided deep image quality assessment
  framework.
\newblock {\em IEEE Transactions on Multimedia}, 19(11):2505--2520, 2017.

\bibitem{8572733}
Mai Xu, Chen Li, Zhenzhong Chen, Zulin Wang, and Zhenyu Guan.
\newblock Assessing visual quality of omnidirectional videos.
\newblock {\em IEEE Transactions on Circuits and Systems for Video Technology},
  29(12):3516--3530, 2019.

\bibitem{qiu2021blind}
Miaomiao Qiu and Feng Shao.
\newblock Blind 360-degree image quality assessment via saliency-guided
  convolution neural network.
\newblock {\em Optik}, 240:166858, 2021.

\bibitem{su2017learning}
Yu-Chuan Su and Kristen Grauman.
\newblock Learning spherical convolution for fast features from 360 imagery.
\newblock {\em Advances in Neural Information Processing Systems}, 30, 2017.

\bibitem{coors2018spherenet}
Benjamin Coors, Alexandru~Paul Condurache, and Andreas Geiger.
\newblock Spherenet: Learning spherical representations for detection and
  classification in omnidirectional images.
\newblock In {\em Proceedings of the European conference on computer vision
  (ECCV)}, pages 518--533, 2018.

\bibitem{esteves2018learning}
Carlos Esteves, Christine Allen-Blanchette, Ameesh Makadia, and Kostas
  Daniilidis.
\newblock Learning so (3) equivariant representations with spherical cnns.
\newblock In {\em Proceedings of the European Conference on Computer Vision
  (ECCV)}, pages 52--68, 2018.

\bibitem{4056759}
Ned Greene.
\newblock Environment mapping and other applications of world projections.
\newblock {\em IEEE Computer Graphics and Applications}, 6(11):21--29, 1986.

\bibitem{chao2018salgan360}
Fang-Yi Chao, Lu~Zhang, Wassim Hamidouche, and Olivier Deforges.
\newblock Salgan360: Visual saliency prediction on 360 degree images with
  generative adversarial networks.
\newblock In {\em Proc. IEEE ICMEW}, 2018.

\bibitem{Pan_2017_SalGAN}
Junting Pan, Cristian Canton, Kevin McGuinness, Noel~E. O'Connor, Jordi Torres,
  Elisa Sayrol, and Xavier~and Giro-i Nieto.
\newblock {SalGAN}: Visual saliency prediction with generative adversarial
  networks.
\newblock In {\em arXiv}, January 2017.

\bibitem{cheng2018cube}
Hsien-Tzu Cheng, Chun-Hung Chao, Jin-Dong Dong, Hao-Kai Wen, Tyng-Luh Liu, and
  Min Sun.
\newblock Cube padding for weakly-supervised saliency prediction in 360 videos.
\newblock In {\em Proc. IEEE/CVF CVPR}, pages 1420--1429, 2018.

\bibitem{9122430}
F.~{Chao}, L.~{Zhang}, W.~{Hamidouche}, and O.~{Deforges}.
\newblock A multi-fov viewport-based visual saliency model using adaptive
  weighting losses for 360-degree images.
\newblock {\em IEEE Transactions on Multimedia}, pages 1--1, 2020.

\bibitem{dahou2021atsal}
Yasser Dahou, Marouane Tliba, Kevin McGuinness, and Noel O’Connor.
\newblock {ATSal}: An attention based architecture for saliency prediction in
  360 videos.
\newblock In {\em International Conference on Pattern Recognition}, pages
  305--320, 2021.

\bibitem{Zhang_2018_ECCV}
Ziheng Zhang, Yanyu Xu, Jingyi Yu, and Shenghua Gao.
\newblock Saliency detection in 360-degree videos.
\newblock In {\em Proc. ECCV}, September 2018.

\bibitem{qiao2020viewport}
Minglang Qiao, Mai Xu, Zulin Wang, and Ali Borji.
\newblock Viewport-dependent saliency prediction in 360 video.
\newblock {\em IEEE Trans. Multimed.}, 23:748--760, 2020.

\bibitem{yun2022panoramic}
Heeseung Yun, Sehun Lee, and Gunhee Kim.
\newblock Panoramic vision transformer for saliency detection in 360-degree
  videos.
\newblock In {\em Computer Vision--ECCV 2022: 17th European Conference, Tel
  Aviv, Israel, October 23--27, 2022, Proceedings, Part XXXV}, pages 422--439.
  Springer, 2022.

\bibitem{Djilali_2021_ICCV}
Yasser Abdelaziz~Dahou Djilali, Tarun Krishna, Kevin McGuinness, and Noel~E.
  O'Connor.
\newblock Rethinking 360deg image visual attention modelling with unsupervised
  learning.
\newblock In {\em Proc. IEEE/CVF ICCV}, pages 15414--15424, October 2021.

\bibitem{eder2020tangent}
Marc Eder, Mykhailo Shvets, John Lim, and Jan-Michael Frahm.
\newblock Tangent images for mitigating spherical distortion.
\newblock In {\em Proc. IEEE/CVF CVPR}, pages 12426--12434, 2020.

\bibitem{gberta_2021_ICML}
Gedas Bertasius, Heng Wang, and Lorenzo Torresani.
\newblock Is space-time attention all you need for video understanding?
\newblock In {\em Proceedings of the International Conference on Machine
  Learning (ICML)}, July 2021.

\bibitem{xu2018gaze}
Yanyu Xu, Yanbing Dong, Junru Wu, Zhengzhong Sun, Zhiru Shi, Jingyi Yu, and
  Shenghua Gao.
\newblock Gaze prediction in dynamic 360 immersive videos.
\newblock In {\em Proc. IEEE/CVF CVPR}, pages 5333--5342, 2018.

\bibitem{xu2018predicting}
Mai Xu, Yuhang Song, Jianyi Wang, MingLang Qiao, Liangyu Huo, and Zulin Wang.
\newblock Predicting head movement in panoramic video: A deep reinforcement
  learning approach.
\newblock {\em IEEE transactions on pattern analysis and machine intelligence},
  2018.

\bibitem{9301766}
F.~Y. {Chao}, C.~{Ozcinar}, L.~{Zhang}, W.~{Hamidouche}, O.~{Deforges}, and
  A.~{Smolic}.
\newblock Towards audio-visual saliency prediction for omnidirectional video
  with spatial audio.
\newblock In {\em 2020 IEEE International Conference on Visual Communications
  and Image Processing (VCIP)}, pages 355--358, 2020.

\bibitem{bylinskii2018different}
Zoya Bylinskii, Tilke Judd, Aude Oliva, Antonio Torralba, and Fr{\'e}do Durand.
\newblock What do different evaluation metrics tell us about saliency models?
\newblock {\em IEEE Trans. Pattern Anal. Mach. Intell.}, 41(3):740--757, 2018.

\bibitem{salfbnet}
Guanqun Ding, Nevrez İmamoğlu, Ali Caglayan, Masahiro Murakawa, and Ryosuke
  Nakamura.
\newblock {SalFBNet}: Learning pseudo-saliency distribution via feedback
  convolutional networks.
\newblock {\em Image and Vision Computing}, 120:104395, 2022.

\bibitem{loshchilov2017decoupled}
Ilya Loshchilov and Frank Hutter.
\newblock Decoupled weight decay regularization.
\newblock {\em arXiv preprint arXiv:1711.05101}, 2017.

\bibitem{morgado2020learning}
Pedro Morgado, Yi~Li, and Nuno Nvasconcelos.
\newblock Learning representations from audio-visual spatial alignment.
\newblock {\em Advances in Neural Information Processing Systems}, 33, 2020.

\bibitem{8578773}
Du~Tran, Heng Wang, Lorenzo Torresani, Jamie Ray, Yann LeCun, and Manohar
  Paluri.
\newblock A closer look at spatiotemporal convolutions for action recognition.
\newblock In {\em 2018 IEEE/CVF Conference on Computer Vision and Pattern
  Recognition}, pages 6450--6459, 2018.

\end{thebibliography}

\end{document}